\documentclass{article}
\usepackage[fontsize=9pt]{scrextend}
\usepackage{spconf,amsmath,graphicx}
\usepackage{xcolor}
\usepackage{hyperref}
\usepackage{multirow}
\usepackage{booktabs}
\usepackage{amssymb}

\title{Show and Speak: Directly Synthesize Spoken Description of Images 
}
\name{Xinsheng Wang$^{1,2,*}$\thanks{* Corresponding author: wangxinsheng@stu.xjtu.edu.cn}, Siyuan Feng$^{2}$, Jihua Zhu$^{1}$, Mark Hasegawa-Johnson$^{3}$, Odette Scharenborg$^{2}$}

\address{$^{1}$School of Software Engineering, Xi’an Jiaotong University, Xi'an, China \\
    $^{2}$Multimedia Computing Group, Delft University of Technology, Delft, The Netherlands \\
    $^{3}$Department of Electrical and Computer Engineering, University of Illinois at Urbana-Champaign, IL, USA}

\begin{document}

%
\maketitle
\begin{abstract}
This paper proposes a new model, referred to as the show and speak (SAS) model that, for the first time, is able to directly synthesize spoken descriptions of images, bypassing the need for any text or phonemes. The basic structure of SAS is an encoder-decoder architecture that takes an image as input and predicts the spectrogram of speech that describes this image. The final speech audio is obtained from the predicted spectrogram via WaveNet. Extensive experiments on the public benchmark database Flickr8k demonstrate that the proposed SAS is able to synthesize natural spoken descriptions for images, indicating that synthesizing spoken descriptions for images while bypassing text and phonemes is feasible.
\end{abstract}


\section{Introduction}
A system that can describe visual scenes in natural language has great potential for helping, for instance, visually-impaired people ``see'' the world. Recent research in this direction is called image captioning \cite{hossain2019comprehensive}, which aims to automatically generate textual descriptions of images. Image captioning systems that automatically generate textual captions of images are inspired by the architecture of neural machine translation and have a neural encoder-decoder structure \cite{vinyals2015show,jia2015guiding}. Recently, benefiting from the development of attention mechanisms \cite{xu2015show,chen2018boosted,li2019entangled,huang2019attention,cornia2020meshed} and training strategies \cite{rennie2017self,chen2017show,keneshloo2019deep}, the task of image captioning has achieved impressive results, making this technology more and more likely to be used in reality. However, nearly half of the world's languages do not have a written form \cite{lewis2015ethnologue}, which means that speakers of those languages cannot benefit from any text-based technologies, including image captioning. In order to make this type of technology available for all languages, it is necessary to develop an image captioning method that bypasses text.

The first work that tried to generate image captions bypassing text is proposed by Hasegawa-Johnson et al. \cite{hasegawa2017image2speech}. In their work, the authors proposed the image-to-speech task, which was based on an intermediate representation of the speech signal in terms of phoneme sequences. Their system first performs an image-to-phoneme generation process, after which the generated phoneme sequence can be used to synthesize the audio signal. Most recently, Van der Hout et al. \cite{van2020evaluating} improved the image-to-phoneme part of the original system \cite{hasegawa2017image2speech} by changing the image encoder structure. Moreover, they investigated how such an image-to-phoneme system could be evaluated objectively by comparing several objective evaluation measures to human ratings. Developing an image-to-phoneme system depends on large amounts of (automatic) alignments of the speech signal with the phonemes. Creating these phoneme annotations requires linguistic expertise. Although Hasegawa-Johnson et al. \cite{hasegawa2017image2speech} also investigated the possibility of using automatically discovered speech units, those units performed subpar in the image-to-phoneme task, and the speech synthesis process based on such speech units was not investigated. Taken together, the image-to-phoneme-to-speech approach is difficult to implement for unwritten languages. 

In order to make an image captioning system able to bypass the dependency on both text and phonemes, this paper presents an image-to-speech generation method which can synthesize spoken descriptions directly from images. The basic architecture of the proposed method is an attention-guided sequence-to-sequence model. Moreover, 
in order to suppress the embedding of image regions that would not be part of a human-generated description,
an embedding constraint is implemented for the image encoder. This model, referred to as the Show and Speak (SAS) model, takes an image as input and outputs the synthesized spoken description of the image\footnote{The synthesized examples, source code, and database can be found at: \href{https://xinshengwang.github.io/projects/SAS/}{https://xinshengwang.github.io/projects/SAS/} \label{ft:project}}.


\section{Model Architecture}
\begin{figure}[t]
    \centering
    \includegraphics[width=\linewidth]{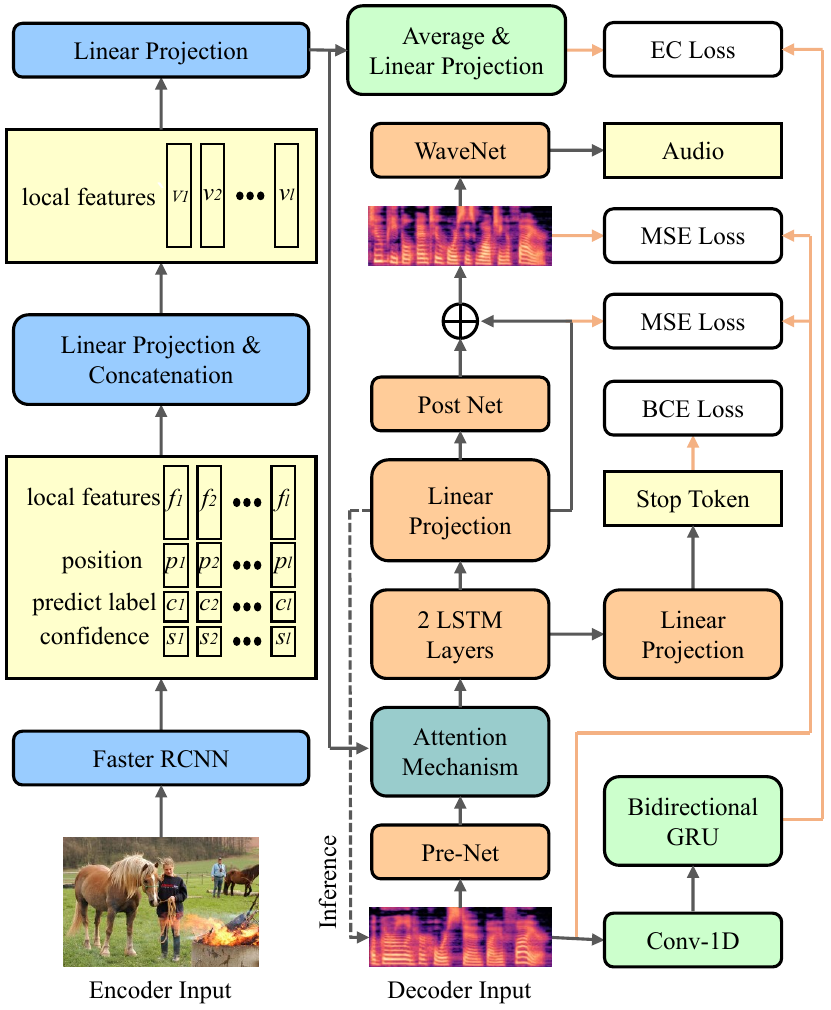}
    \caption{Architecture of the Show and Speak (SAS) model.}
    \label{fig:architecture}
\end{figure}

The proposed SAS model is designed as an encoder-decoder structure. Specifically, the encoder takes an image as input and outputs a sequence of feature vectors of this image. Then, these image feature vectors are taken as input to the decoder which then synthesizes the spectrogram of speech that describes the corresponding image. The architecture of the proposed method is shown in Fig. \ref{fig:architecture} and will be explained in detail below.

\subsection{Encoder}
\label{sc:encoder}
The structure of the encoder is shown in the first column from left in Fig. \ref{fig:architecture}. Given an image, the encoder obtains a sequence of image feature vectors $\{v_1,v_2,...,v_l \}$ of $l$ object regions from the image using a pre-trained object detector. Here, following \cite{anderson2018bottom}, the Faster-RCNN \cite{ren2015faster} model pre-trained on ImageNet \cite{deng2009imagenet} and Visual Genome \cite{krishna2017visual} is adopted to extract image features of $l = 36$ object regions, and these features are referred to as bottom-up features. The extracted feature vectors of one image are presented as $\{f_1,f_2,...,f_l \} \in \mathbb{R}^{l \times d}$, where $ d = 2048$ is the feature dimension. For each local feature $f_i$, the pre-trained Faster-RCNN \cite{anderson2018bottom} provides its position  in the image, predicts the class label, and computes its confidence score (possibility), which are represented as $p_i$, $c_i$, and $s_i$ respectively. Specifically, $p_i \in \mathbb{R}^{5}$ consists of four bounding box coordinate values, i.e., top left $\left( {{\rm{x,y}}} \right)$ and bottom right corner $\left( {{\rm{x,y}}} \right)$, and one ratio value of the bounding box area to the image area. The predicted class label $c_i \in \mathbb{R}^{1601}$ is a one-hot vector, and its corresponding confidence score $s_i$ is a real value. Following \cite{zhou2020unified}, the image feature $v_i$ is obtained via
\begin{equation}
    {v_i} = {f_i} \oplus \left[ {FC\left( {{p_i} \oplus {c_i} \oplus {s_i}} \right)} \right],
\end{equation}
where $\oplus$ means concatenation and FC is the linear projection with 1024 units. Then the image is represented as $V =\{v_1,v_2,...,v_l \} \in \mathbb{R}^{36 \times 3072} $. Finally, in order to create image representations that are more consistent with spoken captions, the image features are passed through two linear transform layers of 1025 and 512 units respectively to get image embeddings with the dimension of 512. The decoder is trained (parameters of the pre-trained Faster-RCNN are fixed) in the encoder-decoder system with the extra embedding constraint that will be introduced in Section \ref{sc:objective_function}.

\subsection{Decoder}
The structure of the decoder is shown in the middle column in Fig. \ref{fig:architecture} (from the decoder input to the spectrogram before the WaveNet). The decoder takes the image feature sequence output from the encoder as input to synthesize speech spectrograms in an autoregressive way. The speech is represented by 80 channel log mel spectrogram computed through a short-time Fourier transform (STFT) with 50 ms frame size and 12.5 ms frame hop. The decoder architecture follows the structure of Tacotron2's decoder \cite{shen2018natural}. Specifically, the generated spectrogram frame from the previous time step passes through a Pre-Net and then is concatenated with an attention context vector before passing through two LSTM layers. The attention context vector is obtained from the encoder output with the location-sensitive attention \cite{chorowski2015attention}, and the Pre-Net is a linear transform layer with 2 fully connected layers both of which have 256 hidden units. The output of the LSTM is concatenated with the attention context vector and then passes through a linear projection to generate the spectrogram frame of the next time step. Then, the generated spectrogram passes through a Post-Net, which consists of 5 convolutional layers with 512 filters, to get an improved spectrogram that is added to the spectrogram before the Post-Net in an element-wise way, achieving the final generated spectrogram. Finally, the generated spectrograms are inverted into time-domain waveform samples via a modified version of WaveNet \cite{oord2016wavenet} in \cite{shen2018natural}.

\subsection{Objective function}
\label{sc:objective_function}
Following the objective function in Tactron2 \cite{shen2018natural},  mean squared error (MSE) is used to optimize the generation of spectrograms before and after Post-Net. Binary cross-entropy (BCE) is used to train the “Stop Token” prediction module that is similar to the module in \cite{shen2018natural}. The stop token prediction allows for the model to dynamically determine the length of the predicted spectrogram sequence instead of synthesizing a fixed-length sequence.
In parallel to the prediction of the spectrograms and stop tokens, an image embedding constraint (EC) loss is introduced to penalize any component in the image embedding that cannot be predicted from the 
spoken caption, i.e., any component of the image embedding that  is semantically independent of the caption. The rounded boxes with green background in Fig. \ref{fig:architecture} show the operations for the image embedding constraint. The image global feature vector is obtained by averaging the encoder outputs, and a linear transform layer is implemented on the averaged vector to get the final image global feature vector that is used to calculate the EC loss. The neural network structure to get the speech embedding vector is similar to the speech encoder in \cite{wang2020s2igan}. Specifically, the ground-truth speech spectrogram first passes through a 1-D convolutional layer, and the fixed-length speech feature vector is obtained by averaging the output of a two-layer bi-directional gated recurrent units (GRU). The matched image-speech vectors should be close to each other, while at the same time different from other unmatched vectors. To that end, we use the Masked Margin Softmax (MMS) method \cite{ilharco2019large} to obtain the EC loss.
We denote losses of spectrogram synthesis, stop token prediction, and image embedding constraint by ${\cal L}_s$, ${\cal L}_{st}$, and ${\cal L}_{ec}$ respectively. The total loss for training the SAS model in an end-to-end way is given by
\begin{equation}
    {\cal L} = {{\cal L}_s} + {{\cal L}_{st}} + \lambda {{\cal L}_{ec}},
\end{equation}
where $\lambda$ is a hyperparameter to balance the image embedding constraint. The value of $\lambda$ is experimentally set as 0.25 out of $\{0.1,0.25,0.5,0.75,1.0\}$.

\section{Experimental setting}
\subsection{Database}
Following the previous experiments on the image-to-phoneme task \cite{hasegawa2017image2speech,van2020evaluating}, Flickr8k \cite{hodosh2013framing} is used in our experiments. This database contains 8,000 images, and each image has 5 textual descriptions. There is a Flickr-Audio database \cite{harwath2015deep} which contains speech recordings for the corresponding textual descriptions. However, these recordings come from more than 100 different speakers, making speech synthesis quite a challenging task. Here, we will not take the multi-speaker speech synthesis problem into consideration, but adopt a text-to-speech (TTS) system \cite{shen2018natural} to synthesize the spoken captions on the basis of textual captions from a single speaker. This TTS system is pre-trained on LJSpeech \cite{ljspeech17} which consists of 13,100 audio clips recorded from a single speaker. We split Flickr8k in the standard way: 6,000 images for training and 1,000 images both for development and test set.

\subsection{Evaluation metrics}
The image-to-speech task is evaluated in terms of how well the synthesized spoken caption describes its corresponding image. However, it is difficult to directly evaluate the spoken captions. In order to objectively measure the performance of our system,  the synthesized speech is automatically transcribed to text. To that end,  an automatic speech recognition (ASR) system\footnote{https://kaldi-asr.org/models/m13} built with  Kaldi \cite{povey2011kaldi} is adopted. The ASR system consists of a hybrid factorized time-delay neural network (TDNN-F) \cite{povey2018semi} acoustic model (AM) and a four-gram language model (LM), both trained using the 960-hour Librispeech English database \cite{panayotov2015librispeech}.

The transcribed textual captions are then evaluated using evaluation metrics for image captioning \cite{huang2019attention,cornia2020meshed}:  BLEU1 (B1), BLEU2 (B2), BLEU3 (B3), BLEU4 (B4), METEOR (M), ROUGE (R), and CIDEr (C). Because the evaluation is performed on the textual captions that are transcribed from the speech captions via the ASR system, higher scores of those metrics can also reflect the good quality of synthesized speech to a certain extent as a worse quality of the synthesized speech would seriously affect the accuracy of the ASR system.

\subsection{Training Details}
\label{sc:training}
We train the SAS network using the Adam optimizer with a warmup in the first 4,000 iterations, and a learning rate that decreases with a continuous exponential decrease from 2e-3.  

The standard neural sequence-to-sequence training procedure, referred to as the teacher-forcing method, feeds the decoder with the ground-truth spectrogram. In the inference stage, this training method could yield errors that can accumulate quickly along the generated sequence due to the discrepancy between training and inference. Here, we adopt the scheduled sampling \cite{bengio2015scheduled} to alleviate this problem. However, we found that when the percent of ground-truth input during the training process decreases to a small value, the generation of speech would be seriously affected. So, we use the inverse sigmoid decay method \cite{bengio2015scheduled} for the percent of ground-truth input with a minimum value of 97.5\%. The effect of this minimum value will be discussed in Section \ref{sc:ablation_study}.

\section{Results}
\subsection{Objective Results}
As the proposed SAS model is the first method that directly synthesizes spoken descriptions of images, there are no existing methods to which it can be compared fairly. Therefore, we take the state-of-the-art image-to-phoneme method \cite{van2020evaluating} to present an upper bound performance on the image-to-speech task. Since this method \cite{van2020evaluating} only generates phonemes rather than speech, we first synthesized the speech based on the generated results of \cite{van2020evaluating}. Specifically, we synthesized the speech with the word sequences generated by \cite{van2020evaluating} (which were used in their human rating study) using the same TTS system that we used to create our ground-truth speech data \cite{shen2018natural}. This topline model is referred to as the phoneme-based method from here onwards.
 
 Note that the image-to-phoneme system \cite{van2020evaluating} is based on phonemes that were obtained using a well-trained same-language ASR, which means that the phoneme-based method cannot be applicable to an unwritten language (see also the Introduction). 

Table \ref{tb:compared_results} shows that the phoneme-based method outperforms our SAS method on all evaluation metrics. The explanation for the worse performance of our SAS model is likely that the end-to-end image-to-speech task is much more challenging than the image-to-phoneme task, due to the following reasons: 1) for a stretch of speech, SAS’s spectrogram sequence is much longer than its transcribed phoneme sequence, and 2) in the image-to-phoneme model, the phoneme generation process during inference can be seen as an autoregressive phoneme prediction process that predicts a phoneme based on an implicitly learned phoneme dictionary at each step. Consequently, it can generate a meaningful phoneme at each step, while there is no dictionary for spectrograms in the SAS model.

\begin{table}[h]
\centering
\vspace{-0.5cm} 
\setlength{\tabcolsep}{1.8mm}
\caption{Compared with the phoneme-based (image-to-phoneme-to-speech) method.}
\begin{tabular}{lccccccc}
\toprule
          & B1 & B2 & B3 & B4 & M & R & C \\ \hline
Phoneme-based & 47.0  & 28.5  & 16.6  & 9.9   & 16.7   & 33.3     & 23.7  \\
SAS       & 29.6 & 14.7 & 7.2 & 3.5 & 11.3 & 23.2 & 8.0   \\ \bottomrule
\label{tb:compared_results}
\end{tabular}
\end{table}

\begin{figure}[htb]
\vspace{-0.4cm} 
    \centering
    \includegraphics[width=\linewidth]{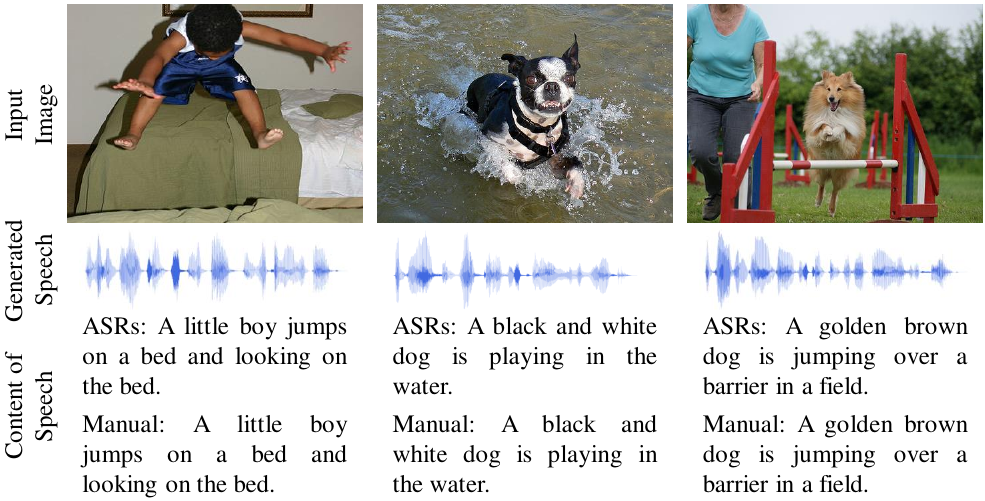}
    \caption{Examples of good synthesized spoken descriptions.}
    \label{fig:good_results}
\end{figure}

\begin{figure}[htb]
\vspace{-0.5cm} 
    \centering
    \includegraphics[width=\linewidth]{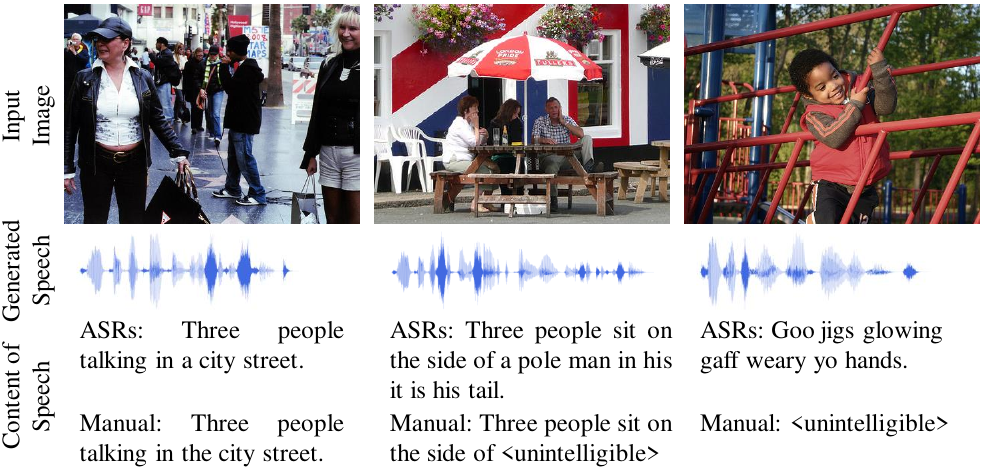}
    \caption{Examples of bad synthesized spoken descriptions. The $\langle$unintelligible$\rangle$ in the manually transcribed text means that the corresponding speech is unintelligible.}
    \label{fig:bad_results}
\end{figure}

\subsection{Subjective Results}

Subjective results that display some good and bad automatically generated spoken captions are shown in Fig. \ref{fig:good_results} and Fig. \ref{fig:bad_results} respectively. To show the speech content, transcribed textual descriptions from the synthesized speech are presented below each image: ``ASRs'' means the textual descriptions are given by the ASR system, and ``Manual'' means the corresponding text is transcribed manually by a human listening to the synthesized speech who does not have access to the corresponding images. The corresponding generated spoken captions and additional examples can be found on the project website\textsuperscript{\ref{ft:project}}.

As shown in Fig. \ref{fig:good_results}, the proposed SAS model can correctly synthesize spoken descriptions for these images, indicating that end-to-end image-to-speech generation bypassing phonemes is feasible. Moreover, based on the fact that spoken captions of low audio quality would yield bad ASR transcriptions, the comparison of the transcriptions provided by the ASR system and those created by the human indirectly show the good quality of the synthesized speech.

However, there are also many cases where our SAS model failed to synthesize good spoken captions. Fig. \ref{fig:bad_results} shows three such cases. In the left image, the synthesized speech is of good quality, i.e., it is intelligible, but the spoken caption does not describe the image well. In the middle image, the quality of the synthesized speech is good at the beginning but gets worse throughout the spoken caption. The right image indicates the worst case in which the synthesized speech is unintelligible. These cases indicate that the robustness of the proposed method needs further improvement.

\subsection{Component analysis}
\label{sc:ablation_study}
Because the image features showed an important impact on the image captioning task \cite{anderson2018bottom}, the performance of the bottom-up features and vanilla ResNet features are compared in this section. Moreover, the proposed image embedding constraint assisting the image embedding is investigated through an ablation study. Finally, the minimum percentage of guided input in the scheduled sampling training process is discussed.

The effect of the image features and the image embedding constraint is shown in Table \ref{tb:ablation_image}. In this table, the Baseline is based on the image features extracted from the pre-trained ResNet-101 rather than the faster-RCNN. Specifically, the image vectors $v_1, v_2,...,v_l$ in Fig. \ref{fig:architecture} are created by scanning the last convolutional layer of ResNet-101 in raster-scan order. In the Baseline, the image embedding constraint is not included. SAS w/o EC means that the SAS model drops the module of image embedding constraint. Compared to the Baseline, in SAS w/o EC, the ResNet-101 features are replaced by the bottom-up features as introduced in Section \ref{sc:encoder}. As shown in this table, the SAS w/o EC shows better performance than the Baseline, indicating the bottom-up features outperform the ResNet-101 features in the image-to-speech task. When the proposed image embedding constraint is added (SAS), the performance increases further, showing the good performance and effectiveness of the image embedding constraint module.

\begin{table}[h]
\vspace{-0.4cm} 
\centering
\setlength{\tabcolsep}{2.2mm}
\caption{The effect of image features and the image embedding constrain on the image-to-speech synthesis. w/o means without.}
\label{tb:ablation_image}
\begin{tabular}{lccccccc}
\toprule
Method   & B1 & B2 & B3 & B4 & M & R & C \\ \hline
Baseline &28.8 & 13.1 & 5.6 & 2.5 & 10.4 & 22.0 &  5.5 \\ 
SAS w/o EC  & 29.6 & 13.9 & 6.3 & 2.8 & 11.1 & 22.8  & 6.8  \\ 
SAS   & \textbf{29.6} & \textbf{14.7} & \textbf{7.2} & \textbf{3.5} & \textbf{11.3} & \textbf{23.2} & \textbf{8.0} \\\bottomrule
\end{tabular}
\end{table}

The effect of the minimum percent $\varepsilon$ of guided sampling (see Section \ref{sc:training}) during the scheduled sampling training process is shown in Table \ref{tb:schudled_sampling}. In this table, $\varepsilon = 100$ means the model is trained fully in a teacher-forcing way, and $\varepsilon = 90$ means the percent of ground-truth input exponentially decreases to 90\% from 100\% during the training. As shown in this table, when the scheduled sampling strategy is implemented ($\varepsilon < 100$), the performance shows obvious changes compared to the fully teacher-forcing training method. Specifically, when the minimum percent of ground-truth input is set as 97.5\%, the SAS model shows the best performance which achieves 29.6\% relative improvement on the BLEU4 compared to the fully-forcing training method. However, when the sampling rate from the real spectrograms (ground-truth input) goes too low, the scheduled sampling leads to a negative effect on the results. Specifically, when $\varepsilon < 95.0$, the generated results become worse than the teacher-forcing training ($\varepsilon = 100$).

\begin{table}[h]
\vspace{-0.4cm} 
\centering
\setlength{\tabcolsep}{2.8mm}
\caption{The effect of minimum guided sampling rate on training the SAS model.}
\label{tb:schudled_sampling}
\begin{tabular}{lccccccc}
\toprule
$\varepsilon$   & B1 & B2 & B3 & B4 & M & R & C \\ \hline
100.0     & 30.0  & 13.7 & 5.8 & 2.7 & 10.8 & 22.8 & 6.7 \\ 
99.0  & \textbf{30.2} & 14.4 & 6.5 & 3.1 & 11.1 & 22.5 & 7.0 \\ 
97.5 & 29.6 & \textbf{14.7} & \textbf{7.2} & \textbf{3.5} & \textbf{11.3} & \textbf{23.2} & \textbf{8.0}  \\ 
95.0  & 28.5 & 13.6 & 6.2 & 3.1 & 10.8 & 22.3 & 7.0 \\ 
92.5 & 27.8 & 13.0 & 6.1 & 2.8 & 10.9 & 21.6 & 6.3 \\ 
90.0   & 25.8 & 12.3 & 5.8 & 2.7 & 10.0 & 20.7 & 6.6  \\ \bottomrule
\end{tabular}
\end{table}

\section{Discussion and Conclusion}
This paper proposes an image-to-speech model, named SAS, which, for the first time, can generate spoken captions of images directly, bypassing any text and phonemes. The proposed SAS model takes an image as input and outputs a spoken description of the image. The results of the image-to-speech experiments show that our SAS model can indeed generate natural spoken descriptions that correctly describe the images. 

Although in the phoneme-based method \cite{hasegawa2017image2speech}, the authors have tried to use automatically discovered speech units as intermediaries (to replace the phonemes), no evidence has shown that these automatically discovered speech units can be used to synthesize natural speech (the speech synthesis stage was not implemented). Moreover, the automatically discovered speech unit-based method showed much worse performance compared to the phoneme-based method on the image-to-speech unit generation task (i.e., without considering the synthesis stage of speech). Performance of the image-to-phone system is better if it uses phonemes transcribed by a well-trained
same-language ASR, but as stated by the authors in \cite{hasegawa2017image2speech}, such a system cannot be used for unwritten languages. The proposed SAS model is the first method that can be used to synthesize natural speech to describe images for unwritten languages, and builds a baseline for this task. 

Compared to the upper bound given by the phoneme-based method \cite{van2020evaluating}, the proposed SAS model still has a large gap to bridge. The SAS model does not always synthesize intelligible speech. An adversarial learning strategy could be considered in the future to improve the quality of the synthesized speech. Finally, the current work is based on the well-resourced English language, and it will be highly interesting to implement this task on a real unwritten language in the future.


\bibliographystyle{IEEEtran}
\begin{small}
\bibliography{refs,strings}
\end{small}

\end{document}